%% file: arxiv.tex
\documentclass[11pt]{article}

\usepackage{iftex}
\ifPDFTeX
  \pdfoutput=1
\fi

\usepackage[dvipsnames]{xcolor}
\usepackage[preprint]{acl}

\usepackage[T1]{fontenc}
\usepackage[utf8]{inputenc}
\usepackage{times}
\usepackage{latexsym}
\usepackage{microtype}
\usepackage{inconsolata}
\usepackage{xurl}

\usepackage{graphicx}
\graphicspath{{figures/}}
\usepackage{booktabs}
\usepackage{multirow}
\usepackage{placeins}
\usepackage{amsmath}
\usepackage{listings}
\definecolor{ForestGreenDark}{HTML}{294A35}
\definecolor{ForestGreen}{HTML}{477A55}
\definecolor{ForestGreenLight}{HTML}{B4D3B9}

\definecolor{TerracottaDark}{HTML}{8C4D3D}
\definecolor{Terracotta}{HTML}{C27E69}
\definecolor{TerracottaLight}{HTML}{E8B9A9}

\definecolor{MutedOchreDark}{HTML}{78602E}
\definecolor{MutedOchre}{HTML}{B18D4C}
\definecolor{MutedOchreLight}{HTML}{E2CC95}

\definecolor{DustyBlueDark}{HTML}{425F79}
\definecolor{DustyBlue}{HTML}{657F9A}
\definecolor{DustyBlueLight}{HTML}{B7C9DD}

\definecolor{LavenderGreyDark}{HTML}{76617F}
\definecolor{LavenderGrey}{HTML}{A995B0}
\definecolor{LavenderGreyLight}{HTML}{D5C2DC}

\usepackage{hyperref}

\hypersetup{
    colorlinks=true,
    linkcolor=MutedOchre,
    citecolor=MutedOchre,
    urlcolor=MutedOchre
}

\title{AnnoCaseLaw: Legal Annotation Benchmark of Civil Negligence}

\author{
  \textbf{Rilton Franzone}$^{1*}$ \quad
  \textbf{Magnus Sesodia}$^{1*}$ \quad
  \textbf{Alina Petrova}$^{2}$ \quad
  \textbf{John Armour}$^{1}$ \\
  \textbf{Thomas Lukasiewicz}$^{1,3}$ \quad
  \textbf{Oana-Maria Camburu}$^{4}$ \quad
  \textbf{Puneet K. Dokania}$^{1}$ \\
  \textbf{Philip Torr}$^{1}$ \quad
  \textbf{Christian Schroeder de Witt}$^{1\dagger}$ \quad
  \textbf{Fabio J. Fehr}$^{1\dagger}$ \\
  \normalfont
  $^{1}$University of Oxford \quad
  $^{2}$Thomson Reuters Labs \\
  \normalfont
  $^{3}$Vienna University of Technology \quad
  $^{4}$University College London
}

\begin{document}
\maketitle
\begingroup
\renewcommand{\thefootnote}{}
\footnotetext{$^{*}$ Equal first authorship;
$^{\dagger}$ equal senior authorship.}
\endgroup

\begin{abstract}
Language models offer a way to scale costly, labour-intensive expert annotation, but their ability to reproduce expert judgments remains uncertain. This is particularly challenging in law, where annotators must distinguish the reasoning supporting judicial decisions from other legally relevant discussion. We present AnnoCaseLaw%
\footnote{
\raisebox{-0.5em}{\includegraphics[height=1.8em]{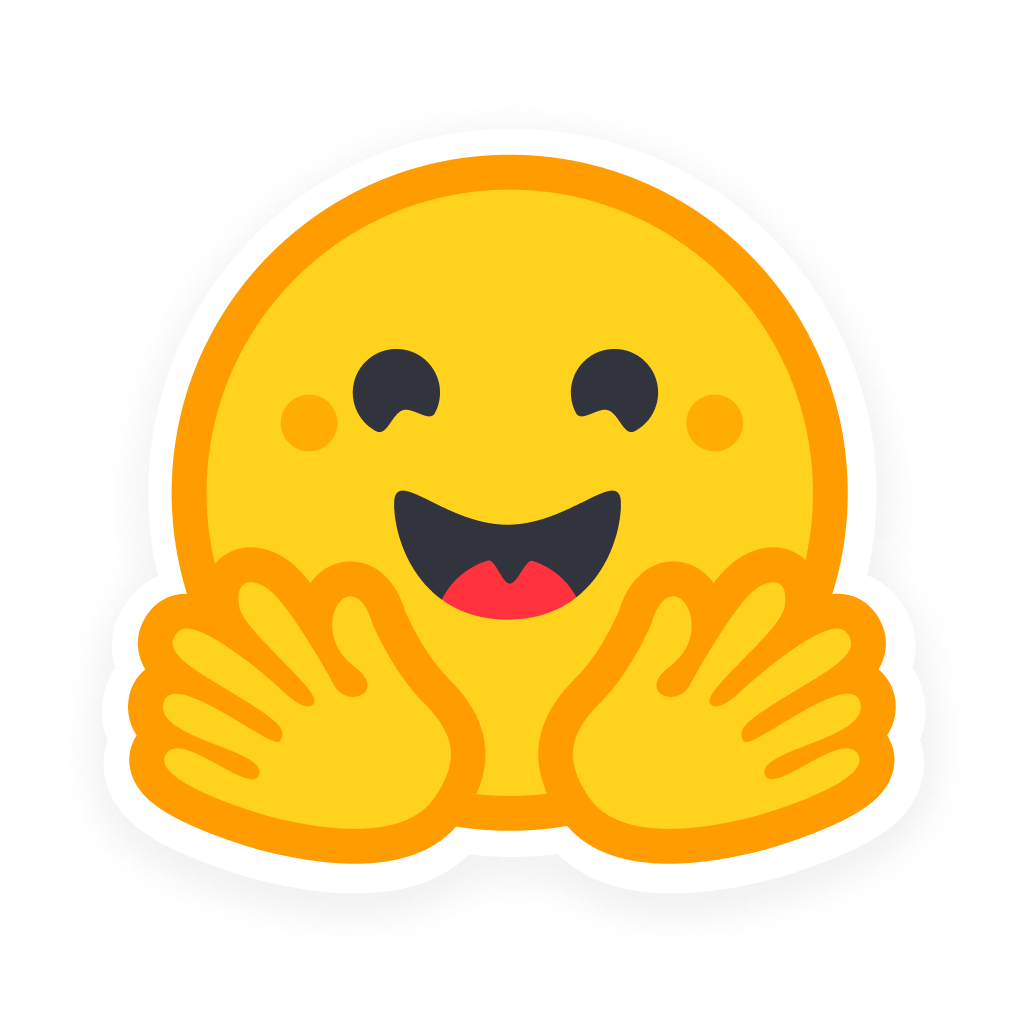}}\ \href{https://huggingface.co/datasets/riltonfranzone/annocaselaw}{Benchmark}
\quad
\raisebox{-0.25em}{\includegraphics[height=1.3em]{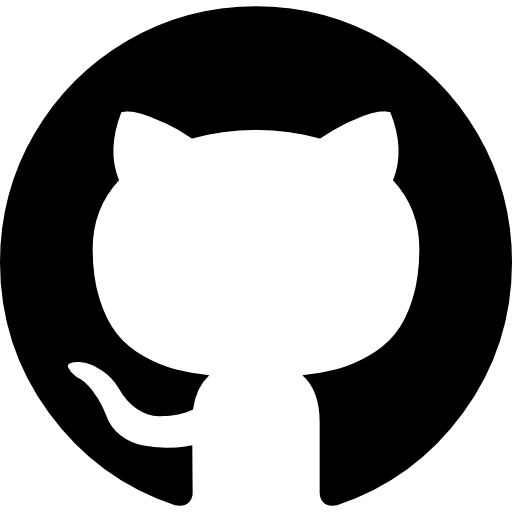}}\ \href{https://github.com/oxai/annocaselaw}{Code}
} as a densely annotated benchmark comprising $471$ U.S. civil-negligence court opinions with more than $12,000$ expert highlights across five annotation categories. We evaluate proprietary frontier, open-weight, legal-specialised, and decision models on their ability to reproduce these annotations. Models perform better on Facts and Outcome than on Application of Law to Facts, Procedural History, and Relevant Precedents. On Application of Law to Facts, proprietary frontier models achieve the strongest balance of reasoning coverage and precision, while the open-weight model Gemma 4 achieves comparable Macro-F1 to proprietary frontier models at a fraction of the estimated inference cost. Our findings highlight the importance of evaluating annotation coverage and cost alongside aggregate scores when assessing the feasibility of automated legal annotation.
\end{abstract}

\section{Introduction}
\label{sec:intro}

\begin{figure}[!t]
  \centering
  \includegraphics[width=\columnwidth]{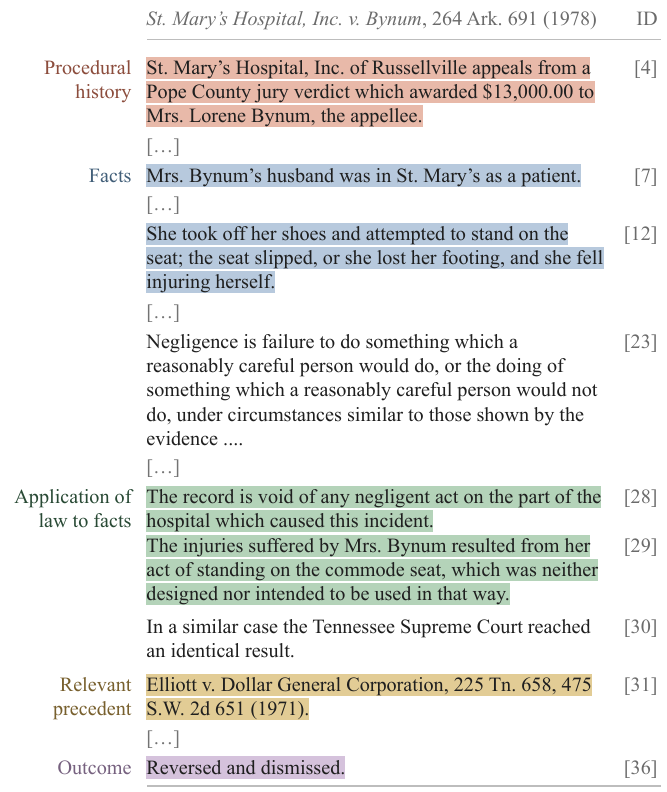}
\caption{\textbf{AnnoCaseLaw expert annotations.}
Five categories illustrated in \textit{St. Mary's Hospital
v. Bynum} (1978). Sentence numbers refer to the full opinion;
bracketed ellipses mark omitted text. Unhighlighted text
was not annotated.}
  \label{fig:annotation-example}
  \vspace{-1em}
\end{figure}

High-quality data annotation is essential for developing and evaluating AI systems, yet remains costly and labour-intensive, particularly in specialised domains. Language models offer a way to scale annotation, with recent studies demonstrating potential cost and efficiency advantages over human annotation \citep{Tan_2024,Parfenova_2025,Gu_2026}. However, models still struggle with legal annotations requiring nuanced understanding, such as identifying complex legal references \citep{Schepers_2025}. Expert legal annotations remain scarce, and some existing datasets rely on keyword-based annotation methods \citep{huang-etal-2024-cmdl,Huang2025}. This raises the question: \textit{can language models reliably reproduce expert legal annotations?}

Identifying the reasoning supporting a judicial decision is particularly demanding, as models may confuse legally worded passages with those that actually support an outcome \citep{Kalamkar_2022,Cheng_2026}. Civil negligence provides a useful setting for studying this problem, as cases with different factual circumstances share a common legal framework, including duty of care and breach. Appellate courts are particularly suitable for studying legal reasoning because they review lower-court decisions against an established factual record, drawing on legal arguments and precedent. Their opinions provide a rich source of evidence on how legal principles are applied to facts. 

We present AnnoCaseLaw (Figure~\ref{fig:annotation-example}), a benchmark comprising $471$ U.S. civil-negligence court opinions with more than $12,000$ expert-annotated highlights. These highlights identify legally significant passages across five categories: Application of Law to Facts (ALF), Procedural History, Relevant Precedents, Facts, and Outcome. Related benchmarks are discussed in Appendix~\ref{app:related}. Our main \textbf{contributions} are:
\begin{itemize}
    \item \textbf{Expert-annotated legal benchmark.} We present AnnoCaseLaw, a densely annotated benchmark capturing legally significant passages, including the application of law to facts and relevant precedents, for systematic evaluation of legal annotation reproduction (Section~\ref{sec:methodology}).
    \item \textbf{Comprehensive model evaluation.} We compare proprietary frontier, open-weight, legal-specialised, and decision models across five annotation categories. Models perform better on Facts and Outcome than on more nuanced categories, including ALF, Procedural History, and Relevant Precedents (Table~\ref{tab:f1}); even frontier models miss about a quarter of expert-marked ALF passages (Figure~\ref{fig:alf-selection}).
    \item \textbf{Annotation quality and cost trade-offs.} We evaluate annotation performance alongside estimated inference cost. Proprietary frontier models achieve the strongest balance of precision and recall on ALF, while Gemma 4 achieves comparable Macro-F1 to proprietary frontier models at a fraction of the estimated inference cost (Figures~\ref{fig:alf-selection} and~\ref{fig:costs}).
\end{itemize}

\section{AnnoCaseLaw Benchmark}
\label{sec:methodology}

\textit{Version note.} An earlier version presented AnnoCaseLaw \citep{sesodia2025annocaselawrichlyannotateddatasetbenchmarking} primarily in the context of legal judgment prediction, with automated case annotation as one of three tasks. This version reframes the work around reproducing expert legal annotations, retaining the underlying corpus and expert annotations while introducing a revised task formulation and new model evaluations and analyses.

\subsection{Dataset Construction}
\label{sec:dataset-construction}
The court opinions were retrieved from the Caselaw Access Project \citep{CAP}. The dataset includes $471$ U.S. civil-negligence cases from three states: Arkansas, Illinois, and New Mexico. Cases were identified from the repository and checked for relevance by the annotators. It includes decisions from state appeals courts, state supreme courts, and the Illinois Claims Court, a specialised tribunal handling negligence claims against the state. Further details on jurisdictional coverage, opinion lengths, and annotation statistics are provided in Appendix~\ref{app:construction}.

\subsection{Expert Annotation Protocol}
\label{sec:annotation-protocol}

The annotations were produced by three legally trained annotators (final-year law students) under the supervision of an experienced legal scholar, following a shared annotation protocol. Each opinion was annotated by one student. The full annotation instructions are provided in Appendix~\ref{app:protocol}.

The protocol distinguishes five categories of passages in judicial opinions, illustrated in Figure~\ref{fig:annotation-example}. \textcolor{ForestGreen}{\textbf{Application of Law to Facts (ALF)}}, \textcolor{Terracotta}{\textbf{Procedural History (PH)}}, and \textcolor{MutedOchre}{\textbf{Relevant Precedents}} capture specific aspects of the court's legal analysis, covering only passages concerning duty of care, breach of duty, and contributory negligence, three central issues in civil negligence law. ALF identifies passages explaining how the court applies the law to the facts, selecting only the reasoning necessary to explain its decision. PH records the proceedings, including earlier courts' decisions and reasons, and the standard the deciding court uses to review them. Relevant Precedents identifies citations to earlier cases and other legal sources directly relevant to the court's reasoning, excluding those used only as background; only cited cases are evaluated in this category (Section~\ref{sec:benchmark-task}). \textcolor{DustyBlue}{\textbf{Facts}} captures events and circumstances before litigation, including facts the court accepts or assumes to be true. Facts are marked comprehensively, including within the court's reasoning. \textcolor{LavenderGreyDark}{\textbf{Outcome}} identifies passages stating the present court's disposition, such as affirming, reversing, remanding, or dismissing.

\input{tables/f1_grouped}

\subsection{Benchmark Task}
\label{sec:benchmark-task}

The benchmark evaluates sentence selection for ALF, PH, Facts
and Outcome, and cited-case selection for Relevant Precedents. For sentence selection, each opinion is split into numbered sentences, and models select sentence numbers for each category. Expert highlights are mapped to the same sentences for evaluation, allowing a sentence to belong to multiple categories. For Relevant Precedents, models receive the opinion text and a list of cited cases extracted automatically using GPT-5.6 Sol, and select those relevant under the annotation protocol. Expert-marked cases missing from the candidate list are outside the recall calculation. Appendix~\ref{app:reference-construction} provides implementation details for sentence alignment and citation-candidate construction.

\vspace{-0.5em}
\subsection{Experimental Setup}
\label{sec:experimental-setup}

We evaluate twelve generative language models and four decision
models, including proprietary frontier, open-weight, and
legal-specialised models. All models receive the full judicial
opinions and instructions adapted from the expert annotation
protocol, without labelled examples. Generative models return
structured JSON, jointly selecting sentences across the four
sentence categories and selecting relevant precedent candidates
in a separate request. Decision models answer a yes/no question
for each sentence--category pair and each precedent candidate;
items are selected when the probability of ``yes'' is at least
0.5. To examine how the question format affects the comparison,
we also evaluate a generative model using yes/no questions on a subset of
opinions (Appendix~\ref{app:yes-no-comparison}). Model settings
and prompts (Appendices~\ref{app:setup}
and~\ref{app:instructions}).

\subsection{Evaluation Metrics}
\label{sec:evaluation-metrics}

We compare model and expert annotations using precision, recall
and F1. Scores are calculated per opinion and averaged
within each category, excluding the opinions without reference
annotations for that category. Macro-F1 is the mean of the five
category F1 scores. We estimate 95\% percentile bootstrap confidence
intervals by resampling opinions; Table~\ref{tab:f1} shows each
score $\pm$ half the interval width as a symmetric approximation.
We also compare the total number of ALF sentences selected by
each model with the expert total, since similar F1 scores can
correspond to different amounts of selected text. Scoring details
are provided in Appendix~\ref{app:scoring}.

\paragraph{Inference Cost.}
We estimate inference costs for all 471 opinions from API token
usage and prices or, for self-hosted models, GPU time and hourly
rental rates. Appendix~\ref{app:costs} details the calculations
and assumptions.

\vspace{2em}
\section{Results}
\label{sec:results}

\begin{figure}[!t]
  \centering
  \includegraphics[width=\columnwidth]{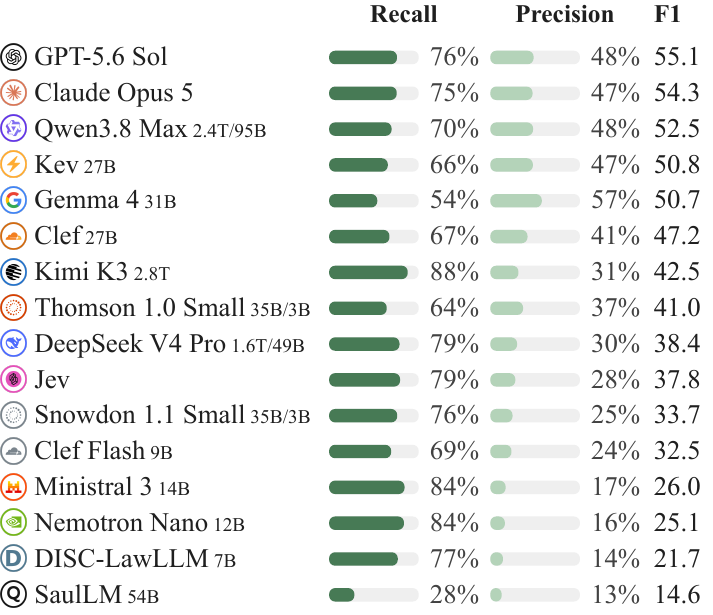}
\caption{\textbf{Proprietary frontier models achieve the
strongest precision--recall balance on ALF.}
Recall, precision and F1 (\%), averaged separately over
opinions and ordered by F1. Mean F1 is not calculated
from the displayed mean precision and recall.}
  \label{fig:alf-selection}
\end{figure}

Table~\ref{tab:f1} shows that models perform better on \textit{Facts} and \textit{Outcome}, with best F1 scores around 80\%, than on \textit{Application of Law to Facts} (ALF), \textit{Procedural History} (PH), and \textit{Relevant Precedents}, where the best scores remain below 65\%.

\newpage

\paragraph{Reasoning Coverage and Precision.}
Within \textit{Application of Law to Facts} (ALF), models trade off reasoning coverage against precision (Figure~\ref{fig:alf-selection}). Proprietary frontier models recover about 75\% of the expert-marked reasoning, while some open-weight models select more text to achieve comparable recall. Gemma achieves competitive F1 through higher precision, but recovers only 54\% of the expert annotations. Appendix~\ref{app:supporting} reports selection counts and additional comparisons.

\begin{figure}[!t]
  \centering
  \includegraphics[width=0.96\columnwidth]{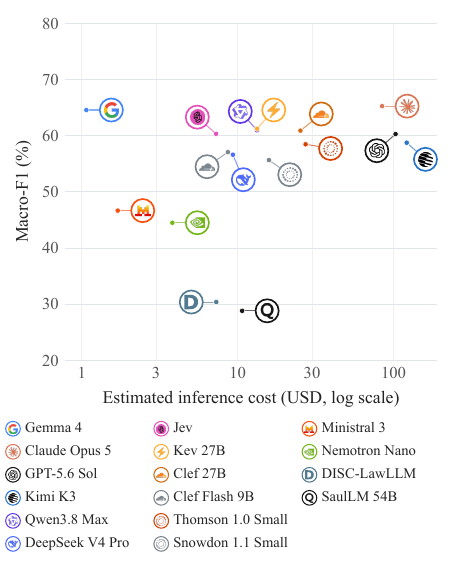}
  \caption{\textbf{Gemma 4 achieves comparable Macro-F1 to proprietary frontier models at lower estimated inference cost.} Macro-F1 across five categories versus estimated inference cost for all 471 opinions. Small dots show values; lines connect them to model labels. Cost calculations are provided in Appendix~\ref{app:costs}.}
  \label{fig:costs}
  \vspace{-1.5em}
\end{figure}

\paragraph{Annotation Performance and Cost.}
These differences in annotation performance also have cost implications (Figure~\ref{fig:costs}). Gemma achieves Macro-F1 close to that of proprietary frontier models at substantially lower inference cost. For the full benchmark, its estimated inference cost is \$1.07, compared with \$84.46 for Claude Opus 5.

\paragraph{Decision models.}
Generative models select annotations jointly, whereas decision
models answer yes/no questions about individual items. To examine
whether this difference affects their comparison, we also test
Gemma with yes/no questions on a subset of 40 opinions. On this sample, Jev,
Kev and Clef achieve higher Macro-F1 than Gemma's yes/no version,
but all decision models remain below its joint selection
(Appendix~\ref{app:yes-no-comparison}), suggesting that the
question format contributes to the observed differences between
models.




\section{Discussion}
\label{sec:discussion}

Current models struggle to reproduce expert legal annotations, particularly the precise reasoning supporting judicial decisions. Models that recover more expert-marked reasoning often select additional text, while more precise models may miss important passages, making aggregate F1 an incomplete measure of annotation quality. Model choice also involves a cost trade-off: smaller open-weight models such as Gemma offer competitive overall performance at substantially lower inference cost, but may recover less legal reasoning. The appropriate model therefore depends on the annotation task, the relative importance of missed and unnecessary selections, and the available budget. These findings suggest that models should be evaluated as potential aids to expert annotation rather than assumed replacements for expert judgment.

Thomson Small improves F1 for reasoning annotation by 7.3 points over its parent, Snowdon Small, gaining precision at the cost of some recall (Appendix~\ref{app:checkpoint-retention}). This suggests that specialisation can improve annotation performance, but since Thomson's training also includes non-legal material, we cannot attribute the improvements to legal adaptation alone.

\section{Conclusion}
\label{sec:conclusion}

AnnoCaseLaw provides a benchmark for evaluating the reproduction of legally significant annotations in U.S. civil-negligence opinions (Section~\ref{sec:methodology}). Frontier models perform better on Facts and Outcome than on more nuanced categories, particularly Application of Law to Facts and Relevant Precedents (Table~\ref{tab:f1}). These findings highlight the need to assess annotation coverage alongside estimated inference cost (Figures~\ref{fig:alf-selection} and~\ref{fig:costs}).



\section*{Limitations}

AnnoCaseLaw provides a densely annotated benchmark of U.S. civil-negligence opinions spanning three states and multiple court types. However, we do not claim that these results will generalise to other jurisdictions, areas of law, annotation protocols, or languages. As the underlying opinions are publicly available, we cannot rule out that models encountered them during training, potentially confounding model comparisons. Thomson 1.0 Small also received training beyond the legal domain, including tax and journalism material, as well as further post-training, so its improvement over Snowdon 1.1 Small cannot be attributed to legal training alone. Decision and generative models used different inference settings, and some generative models required more retries than others. Finally, three final-year law students produced the annotations under the supervision of an experienced legal scholar and following a shared, documented protocol. Each opinion was annotated by one student. Given the scope of this paper, independently annotating opinions to assess inter-annotator agreement was beyond the scope of this study; this remains an important direction for future work.

\section*{Future Work}
Our results highlight the promise of language models for annotation in high-stakes domains such as law. Promising directions for further developing AnnoCaseLaw include fine-tuning smaller open-weight and legal-specialised models, including decision models, to improve annotation coverage while maintaining precision and cost efficiency. Further work should also evaluate whether model-assisted annotation reduces human effort compared with annotation from scratch.

\section*{Ethical considerations}

AnnoCaseLaw contains people’s names and details of their injuries and disputes. These details can be sensitive even though the court opinions are public. The goal of this dataset is to enable research on legal annotation, and performance on this benchmark does not establish whether a model can provide reliable legal advice or make legal decisions.

\section*{Acknowledgments}

This work is supported by the UKRI grant: Turing AI Fellowship EP/W002981/1. CSDW also acknowledges funding from the UK Government Grant ICRF2425-8-160, the Cooperative AI Foundation, Armasuisse Science+Technology, an OpenAI Superalignment Fast Grant, and Schmidt Futures. We extend our gratitude to Alexandru \mbox{Circiumaru}, Ayban Elliott-Renhard, and Aleksandra Kobyasheva for their invaluable contributions as domain expert annotators.

\bibliography{main}

\appendix
\input{appendix}

\end{document}

%% file: tables/f1_grouped.tex
\begin{table*}[t!]
\centering
\small
\setlength{\tabcolsep}{4.5pt}
\resizebox{\textwidth}{!}{%
\begin{tabular}{llr@{\,}lr@{\,}lr@{\,}lr@{\,}lr@{\,}lr@{\,}l}
\toprule
Group & Model & \multicolumn{2}{c}{ALF} & \multicolumn{2}{c}{PH} & \multicolumn{2}{c}{Precedents} & \multicolumn{2}{c}{Facts} & \multicolumn{2}{c}{Outcome} & \multicolumn{2}{c}{Macro-F1} \\
\midrule
\multirow{2}{*}{Proprietary frontier} & \raisebox{-0.15em}{\includegraphics[height=0.9em]{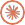}}\hspace{0.35em}Claude Opus 5 & \underline{54.3} & {\scriptsize$\pm$2.1} & 53.1 & {\scriptsize$\pm$2.1} & \textbf{62.8} & {\scriptsize$\pm$2.8} & \textbf{79.2} & {\scriptsize$\pm$1.6} & 77.1 & {\scriptsize$\pm$1.9} & \textbf{65.3} & {\scriptsize$\pm$1.4} \\
 & \raisebox{-0.15em}{\includegraphics[height=0.9em]{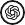}}\hspace{0.35em}GPT-5.6 Sol & \textbf{55.1} & {\scriptsize$\pm$2.2} & 40.8 & {\scriptsize$\pm$2.0} & 61.7 & {\scriptsize$\pm$2.9} & 71.0 & {\scriptsize$\pm$1.7} & 72.9 & {\scriptsize$\pm$2.2} & 60.3 & {\scriptsize$\pm$1.4} \\
\midrule
\multirow{3}{*}{Open, $>$1T} & \raisebox{-0.15em}{\includegraphics[height=0.9em]{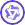}}\hspace{0.35em}Qwen3.8 Max {\scriptsize\textcolor{gray}{2.4T/95B}} & 52.5 & {\scriptsize$\pm$2.1} & 48.0 & {\scriptsize$\pm$2.1} & 58.0 & {\scriptsize$\pm$3.0} & 69.3 & {\scriptsize$\pm$1.7} & 77.2 & {\scriptsize$\pm$2.1} & 61.0 & {\scriptsize$\pm$1.3} \\
 & \raisebox{-0.15em}{\includegraphics[height=0.9em]{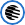}}\hspace{0.35em}Kimi K3 {\scriptsize\textcolor{gray}{2.8T}} & 42.5 & {\scriptsize$\pm$1.9} & 46.1 & {\scriptsize$\pm$2.1} & \underline{61.9} & {\scriptsize$\pm$2.8} & 73.8 & {\scriptsize$\pm$1.8} & 69.5 & {\scriptsize$\pm$2.2} & 58.8 & {\scriptsize$\pm$1.4} \\
 & \raisebox{-0.15em}{\includegraphics[height=0.9em]{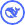}}\hspace{0.35em}DeepSeek V4 Pro {\scriptsize\textcolor{gray}{1.6T/49B}} & 38.4 & {\scriptsize$\pm$1.9} & 41.0 & {\scriptsize$\pm$2.3} & 54.8 & {\scriptsize$\pm$3.0} & 69.7 & {\scriptsize$\pm$2.0} & 79.3 & {\scriptsize$\pm$2.1} & 56.6 & {\scriptsize$\pm$1.4} \\
\midrule
\multirow{4}{*}{Open, 31--54B} & \raisebox{-0.15em}{\includegraphics[height=0.9em]{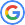}}\hspace{0.35em}Gemma 4 {\scriptsize\textcolor{gray}{31B}} & 50.7 & {\scriptsize$\pm$2.0} & \textbf{56.9} & {\scriptsize$\pm$2.2} & 60.6 & {\scriptsize$\pm$2.9} & 74.8 & {\scriptsize$\pm$1.7} & \underline{79.9} & {\scriptsize$\pm$1.9} & \underline{64.6} & {\scriptsize$\pm$1.2} \\
 & \raisebox{-0.15em}{\includegraphics[height=0.9em]{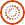}}\hspace{0.35em}Thomson 1.0 Small\smash{$^\dagger$} {\scriptsize\textcolor{gray}{35B/3B}} & 41.0 & {\scriptsize$\pm$2.0} & 51.7 & {\scriptsize$\pm$2.7} & 58.5 & {\scriptsize$\pm$3.3} & 74.9 & {\scriptsize$\pm$1.7} & 66.5 & {\scriptsize$\pm$2.7} & 58.5 & {\scriptsize$\pm$1.5} \\
 & \raisebox{-0.15em}{\includegraphics[height=0.9em]{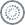}}\hspace{0.35em}Snowdon 1.1 Small {\scriptsize\textcolor{gray}{35B/3B}} & 33.7 & {\scriptsize$\pm$1.8} & 49.5 & {\scriptsize$\pm$2.4} & 59.6 & {\scriptsize$\pm$3.1} & 72.8 & {\scriptsize$\pm$1.8} & 62.7 & {\scriptsize$\pm$2.8} & 55.7 & {\scriptsize$\pm$1.5} \\
 & \raisebox{-0.15em}{\includegraphics[height=0.9em]{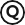}}\hspace{0.35em}SaulLM\smash{$^\dagger$} {\scriptsize\textcolor{gray}{54B}} & 14.6 & {\scriptsize$\pm$1.6} & 20.9 & {\scriptsize$\pm$2.3} & 42.3 & {\scriptsize$\pm$3.0} & 52.9 & {\scriptsize$\pm$2.4} & 13.2 & {\scriptsize$\pm$2.5} & 28.8 & {\scriptsize$\pm$1.4} \\
\midrule
\multirow{3}{*}{Open, $\leq$14B} & \raisebox{-0.15em}{\includegraphics[height=0.9em]{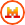}}\hspace{0.35em}Ministral 3 {\scriptsize\textcolor{gray}{14B}} & 26.0 & {\scriptsize$\pm$1.5} & 22.2 & {\scriptsize$\pm$1.7} & 55.3 & {\scriptsize$\pm$2.6} & 59.1 & {\scriptsize$\pm$2.3} & 70.7 & {\scriptsize$\pm$2.3} & 46.7 & {\scriptsize$\pm$1.3} \\
 & \raisebox{-0.15em}{\includegraphics[height=0.9em]{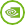}}\hspace{0.35em}Nemotron Nano {\scriptsize\textcolor{gray}{12B}} & 25.1 & {\scriptsize$\pm$1.4} & 31.3 & {\scriptsize$\pm$2.5} & 52.9 & {\scriptsize$\pm$2.9} & 60.6 & {\scriptsize$\pm$2.4} & 52.6 & {\scriptsize$\pm$2.7} & 44.5 & {\scriptsize$\pm$1.4} \\
 & \raisebox{-0.15em}{\includegraphics[height=0.9em]{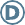}}\hspace{0.35em}DISC-LawLLM\smash{$^\dagger$} {\scriptsize\textcolor{gray}{7B}} & 21.7 & {\scriptsize$\pm$1.6} & 11.2 & {\scriptsize$\pm$1.4} & 46.9 & {\scriptsize$\pm$2.6} & 30.4 & {\scriptsize$\pm$2.0} & 41.7 & {\scriptsize$\pm$3.2} & 30.4 & {\scriptsize$\pm$1.4} \\
\midrule
\multirow{4}{*}{Decision models} & \raisebox{-0.15em}{\includegraphics[height=0.9em]{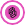}}\hspace{0.35em}Jev \scriptsize{(Proprietary)} & 37.8 & {\scriptsize$\pm$1.8} & \underline{54.0} & {\scriptsize$\pm$2.2} & 59.4 & {\scriptsize$\pm$2.9} & 73.6 & {\scriptsize$\pm$1.6} & 77.0 & {\scriptsize$\pm$2.0} & 60.4 & {\scriptsize$\pm$1.3} \\
 & \raisebox{-0.15em}{\includegraphics[height=0.9em]{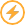}}\hspace{0.35em}Kev {\scriptsize\textcolor{gray}{27B}} \scriptsize{(Open)} & 50.8 & {\scriptsize$\pm$1.9} & 39.7 & {\scriptsize$\pm$2.6} & 61.3 & {\scriptsize$\pm$2.9} & 74.4 & {\scriptsize$\pm$1.6} & \textbf{80.1} & {\scriptsize$\pm$1.9} & 61.3 & {\scriptsize$\pm$1.3} \\
 & \raisebox{-0.15em}{\includegraphics[height=0.9em]{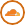}}\hspace{0.35em}Clef {\scriptsize\textcolor{gray}{27B}} \scriptsize{(Open)} & 47.2 & {\scriptsize$\pm$1.8} & 46.9 & {\scriptsize$\pm$2.1} & 58.6 & {\scriptsize$\pm$2.8} & \underline{76.7} & {\scriptsize$\pm$1.6} & 75.1 & {\scriptsize$\pm$2.0} & 60.9 & {\scriptsize$\pm$1.3} \\
 & \raisebox{-0.15em}{\includegraphics[height=0.9em]{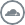}}\hspace{0.35em}Clef Flash {\scriptsize\textcolor{gray}{9B}} \scriptsize{(Open)} & 32.5 & {\scriptsize$\pm$1.4} & 50.3 & {\scriptsize$\pm$2.1} & 55.1 & {\scriptsize$\pm$2.5} & 74.2 & {\scriptsize$\pm$1.7} & 73.3 & {\scriptsize$\pm$2.2} & 57.1 & {\scriptsize$\pm$1.2} \\
\midrule
Baseline & \textit{Select all} & \textit{16.9} &  & \textit{8.3} &  & \textit{48.2} &  & \textit{29.9} &  & \textit{4.0} &  & \textit{21.5} &  \\
\bottomrule
\end{tabular}%
}
\caption{\textbf{Models perform better on Facts and Outcome than on nuanced legal categories.} Per-opinion F1 (\%); $\pm$ denotes the half-width of a 95\% bootstrap confidence interval. Macro-F1 averages the five category scores. The select-all baseline labels every scorable item. Sizes indicate total/active parameters; $\dagger$ marks legal-specialised models. Bold and underline mark the highest and second-highest point estimates per column.}
\label{tab:f1}
\vspace{-1em}
\end{table*}

%% file: appendix.tex
\raggedbottom

\section{Additional related work}
\label{app:related}

Existing legal annotation datasets ask annotators to identify
different parts of a judgment. \citet{Kalamkar_2022} and SCOTUS-Law
\citep{Belfathi_2026} label sentences by their rhetorical role,
including analysis and references to legal authorities. PredEx
contains expert-selected passages supporting judgment outcomes
\citep{Nigam_2024}, while RaVE provides rationales from two experts
and examines disagreements between them \citep{Xu_2023}.
AnnoCaseLaw evaluates selections across five categories, with ALF,
PH and Relevant Precedents restricted to duty of care, breach of
duty and contributory negligence.

\section{Dataset construction}
\label{app:construction}

\subsection{Collection and annotation}
\label{app:annotators}

The opinions in AnnoCaseLaw were retrieved from
the Caselaw Access Project \citep{CAP}. Collection focused on
Illinois, Arkansas and New Mexico, whose case reports were available
for bulk download. Three final-year law students searched for terms
such as \textit{negligence}, \textit{breach of duty} and
\textit{duty of care}, then checked whether each opinion concerned
civil negligence.

Each opinion was annotated by one student following the written
protocol in Appendix~\ref{app:protocol}. An experienced legal
scholar supervised the work, with weekly meetings to discuss the
annotations and clarify the instructions.

\subsection{Dataset statistics}
\label{app:dataset-statistics}

The 471 opinions contain 12,538 original highlights. They come from
New Mexico (198), Arkansas (152), Illinois (119), Arizona (one) and
Alaska (one). After cleaning, opinions contain a median of 116
sentence units and 2,993 words, with ranges of 25--474 sentences and
543--13,278 words.

Table~\ref{tab:eligible} summarises the annotations used for evaluation.
Of the 61,847 sentence units, 472 are excluded because their labels
cannot be assigned reliably under the alignment rules below.
Precedent selection uses 7,694 candidate cases, of which 2,139 match
the expert annotations.

\begin{table}[t]
\centering\small
\setlength{\tabcolsep}{4pt}
\begin{tabular}{lrrr}
\toprule
Category & Opinions & Spans & Items \\
\midrule
Facts & 470 & 3,906 & 10,848 \\
PH & 448 & 1,601 & 2,254 \\
ALF & 471 & 2,770 & 5,164 \\
Outcome & 462 & 927 & 1,015 \\
Precedents & 409 & 3,334 & 2,139 \\
\bottomrule
\end{tabular}
\caption{Annotation counts by category. Opinions are those with reference items; spans are original highlights. Items are distinct cited cases for Precedents and sentences for the other categories.}
\label{tab:eligible}
\end{table}

\subsection{Reference construction}
\label{app:reference-construction}

\paragraph{Sentence labels.}
We remove annotator summaries, appended lower-court opinions and
separate concurring or dissenting opinions from the input text.
A deterministic splitter uses punctuation and paragraph boundaries,
with rules for legal abbreviations, citations and footnotes.
Expert-highlighted excerpts are matched to the cleaned text,
allowing for differences in whitespace, page markers and OCR errors.
Repeated excerpts were aligned using the original annotated documents
or, for 46 highlights, agreement between two language models. The
original annotation categories were retained.

A sentence receives a category when highlights in that category cover
at least half of its non-whitespace characters. Shorter highlights
also qualify when at least half of one highlight lies within the
sentence, with an overlap of at least 15 non-whitespace characters.
For Outcome, the minimum is six characters to retain short decisions
such as ``Affirmed.'' Categories are assigned independently, so a
sentence can receive more than one label.

Of the 9,204 highlights across the four sentence categories, 251 are
omitted from alignment because they are very short, occur frequently,
fall outside the retained text or cannot be located.
Separately, 472 sentence units with highlights that meet neither rule,
or whose labels depend on unresolved repeated excerpts, are excluded
from scoring for every model and all four sentence categories. They
remain in the model input without exclusion markers. Unhighlighted
sentences remain available for evaluation.

\paragraph{Precedent candidates.}
A separate GPT-5.6 Sol pass, with medium reasoning effort, extracts
case names and citations from each cleaned opinion. The mentions
are checked against the text, allowing for OCR errors, and grouped
by case identity. Candidates are aligned with expert highlights to
identify the reference selections. Every model receives the same
candidate list without reference labels.

This task covers cited cases only; statutes and other legal sources
included in the original annotations are not evaluated. Scores
measure relevance selection from the supplied candidates, so an
expert-marked case missing from the candidate list is outside the
recall calculation.

\section{Experimental details}
\label{app:setup}

\subsection{Generative models}

Tables~\ref{tab:model-roster} and~\ref{tab:settings} give the model
identifiers and inference settings. The twelve generative models
were run in September 2026. Eight used DigitalOcean's inference API;
the four self-hosted models used vLLM with BF16 weights.

Each task uses a separate request containing the full cleaned opinion
and the relevant prompt from Appendix~\ref{app:prompts}.

DISC-LawLLM was developed for Chinese law and uses its authors'
Chinese legal-assistant system message, with a grammar enforcing
the requested ID format.
Nemotron receives an additional instruction to return JSON without
tool calls.

We retry failed or invalid responses with unchanged prompts, retain
the first valid answer and preserve selected IDs during formatting
repairs.

We use temperature 0 for Gemma, Ministral, Qwen and DeepSeek.
For Kimi, Nemotron and DISC-LawLLM, we follow published sampling
settings, while Thomson, Snowdon and SaulLM use checkpoint or
library defaults. Sol and Opus retain provider-default sampling
with the reasoning settings shown in Table~\ref{tab:settings}.

\begin{table*}[t]
\centering\small
\setlength{\tabcolsep}{4pt}
\begin{tabular}{lrrrrrrr}
\toprule
Model & Serving & $T$ & $p$ & $k$ & Thinking & Output cap & Context \\
\midrule
GPT-5.6 Sol & API & -- & -- & -- & high & 32,768 & 1,050,000 \\
Claude Opus 5 & API & -- & -- & -- & adaptive/high & 32,768 & 1,000,000 \\
Qwen3.8 Max & API & 0 & -- & -- & -- & 32,768 & 1,000,000 \\
Gemma 4 31B & API & 0 & -- & -- & -- & 8,192 & 256,000 \\
DeepSeek V4 Pro & API & 0 & -- & -- & -- & 32,768 & 1,048,576 \\
Kimi K3 & API & 1 & -- & -- & high & 32,768 & 1,000,000 \\
Thomson 1.0 Small & vLLM & 1.0 & 0.95 & 20 & on & 65,536 & 98,304 \\
Snowdon 1.1 Small & vLLM & 1.0 & 0.95 & 20 & on & 65,536 & 98,304 \\
SaulLM 54B & vLLM & 1.0 & 1.0 & 50 & -- & 16,384 & 32,768 \\
DISC-LawLLM 7B & vLLM & 0.1 & 0.9 & 50 & -- & 16,384 & 32,768 \\
Ministral 3 14B & API & 0 & -- & -- & -- & 8,192 & 262,144 \\
Nemotron Nano 12B V2 VL & API & 0.6 & 0.95 & -- & -- & 16,384 & 128,000 \\
\bottomrule
\end{tabular}
\caption{Generation settings. Dashes denote no explicit override. Output caps and context sizes are in tokens. SaulLM and DISC-LawLLM use smaller output caps when needed to fit the full opinion; inputs are never truncated.}
\label{tab:settings}
\end{table*}

\begin{table*}[t]
\centering\small
\setlength{\tabcolsep}{4pt}
\begin{tabular}{lp{0.32\textwidth}p{0.26\textwidth}p{0.19\textwidth}}
\toprule
Model & Endpoint or checkpoint & Weights revision & Source \\
\midrule
Sol & \nolinkurl{openai-gpt-5.6-sol} & -- & \citep{Sol_2026} \\
Opus & \nolinkurl{anthropic-claude-opus-5} & -- & \citep{Opus_2026} \\
Qwen & \nolinkurl{qwen3.8-max} & -- & \citep{Qwen_2026} \\
Gemma & \nolinkurl{gemma-4-31B-it} & -- & \citep{Gemma_2026} \\
DeepSeek & \nolinkurl{deepseek-v4-pro} & -- & \citep{DeepSeek_2026} \\
Kimi & \nolinkurl{kimi-k3} & -- & \citep{Kimi_2026} \\
Thomson & \href{https://huggingface.co/thomsonreuters/Thomson-1.0-Small/tree/6f58dd819061240192285228e8c0ffcffd0fc665}{\nolinkurl{thomsonreuters/Thomson-1.0-Small}} & \nolinkurl{6f58dd819061240192285228e8c0ffcffd0fc665} & \citep{ThomsonReport_2026} \\
Snowdon & \href{https://huggingface.co/tri-fair-lab/Snowdon1.1-Small/tree/3e3936f24671a444516a41cb9dbb10cc71e09c9d}{\nolinkurl{tri-fair-lab/Snowdon1.1-Small}} & \nolinkurl{3e3936f24671a444516a41cb9dbb10cc71e09c9d} & \citep{Snowdon_2026} \\
SaulLM & \href{https://huggingface.co/Equall/SaulLM-54B-Instruct/tree/cb41c48a180a0a4c0d18a20808c395d0485c741f}{\nolinkurl{Equall/SaulLM-54B-Instruct}} & \nolinkurl{cb41c48a180a0a4c0d18a20808c395d0485c741f} & \citep{Saul_2024} \\
DISC-LawLLM & \href{https://huggingface.co/ShengbinYue/LawLLM-7B/tree/abc510f9f5f8be2676c13f7786c78d775496f857}{\nolinkurl{ShengbinYue/LawLLM-7B}} & \nolinkurl{abc510f9f5f8be2676c13f7786c78d775496f857} & \citep{DISC_2023} \\
Ministral & \nolinkurl{mistral-3-14B} & -- & \citep{Ministral_2025} \\
Nemotron & \nolinkurl{nemotron-nano-12b-v2-vl} & -- & \citep{Nemotron_2025} \\
\bottomrule
\end{tabular}
\caption{Model identifiers and references. Checkpoint links point to the evaluated revisions, fixing the weights and tokenizer; API aliases do not expose fixed revisions.}
\label{tab:model-roster}
\end{table*}

\subsection{Decision models}
\label{app:decision-models}

We evaluate Jev~\citep{Jev_2026}, Kev~\citep{Kev_2026},
Clef 27B~\citep{Clef_2026} and Clef Flash 9B~\citep{ClefFlash_2026}.
Each model returns a probability for a yes/no question for each
sentence and category, and for each precedent candidate. They receive
the full cleaned opinion and the same substantive annotation rules,
using the question templates in Appendix~\ref{app:decision-questions}.
An item is selected when the returned probability is at least 0.5.

The decision models were run in October 2026. Jev uses DigitalOcean's
\href{https://docs.typesafe.ai/introduction}{\nolinkurl{typesafe-jev-1.13.0}} endpoint. Kev uses
\href{https://huggingface.co/jaredpalmer/kev-27b/tree/af0e6d551bdc}{\nolinkurl{jaredpalmer/kev-27b}}, with checkpoint revision
\texttt{af0e6d551bdc} and code revision \texttt{5e42a7a03f28},
served in BF16 on one B300. Its published score temperature
(approximately 1.3195) is used without refitting.

Clef and Clef Flash use the published
\href{https://huggingface.co/Cloudflare/clef/tree/0b331204bb13fbd2ca93a64df1956f5b55478ce5}{\nolinkurl{Cloudflare/clef}}
and
\href{https://huggingface.co/Cloudflare/clef-flash/tree/8b2e5fd17c09fd49bc2880b805d5323ec7ff4ff3}{\nolinkurl{Cloudflare/clef-flash}}
checkpoints, respectively. Both run in BF16 using the authors'
inference code and published calibration without refitting, on A40
and H100 GPUs. They receive the same opinions, questions and question
groups as Jev, with no input truncation; Clef scores the questions
within each group jointly.

\subsection{Scoring and confidence intervals}
\label{app:scoring}

Precision, recall and F1 compare model selections with the reference
items after excluded sentence IDs are removed. Empty selections
receive zero for all three metrics when the opinion has reference
items for that category. Each metric is averaged over opinions with
at least one reference item, giving them equal weight. Macro-F1 is
the mean of the five category-level F1 scores.

For ALF, we also divide the total number of model selections by the
total number of expert selections across the same opinions:
$\sum_i |S_i|/\sum_i |G_i|$, where $S_i$ and $G_i$ are the model and
reference sets for opinion $i$.

We estimate 95\% percentile confidence intervals using 10,000
bootstrap resamples of opinions \citep{Efron_1993}, using the same
draws for all models. For Macro-F1, each draw samples
from the union of eligible opinions, and each category is averaged
over its eligible sampled opinions. Model differences are calculated
within each draw. Table~\ref{tab:f1} displays the score with $\pm$
half the interval's width, a symmetric approximation to the percentile
interval.

These intervals describe variation across opinions with the
annotations and model outputs fixed. They do not measure variation
across repeated inference runs and are not adjusted for multiple
comparisons.

\subsection{Inference costs}
\label{app:compute}
\label{app:costs}

Costs cover annotation of all 471 opinions across the five categories.
For API models, we multiply recorded token usage by the prices in
Table~\ref{tab:api-costs}. Sol and Opus use prices recorded on
20 September 2026, Jev uses those from 7 October, and the other API
runs use DigitalOcean's 8 October prices. Recorded retry and reasoning
tokens are included, and cache discounts apply only to recorded
cached tokens. Failed requests with unreported usage are not included.

\begin{samepage}
For Thomson, Snowdon, SaulLM, DISC-LawLLM and Kev, we multiply
recorded GPU rental time by the
hourly rate in Table~\ref{tab:gpu-costs}. Rental time includes setup,
retries and idle time; DISC-LawLLM's allocation also covers generations
not used in the reported results. Thomson's estimate is a lower bound
because its recorded duration ends when the run finishes.

For Clef and Clef Flash, we estimate the cost of running all opinions
on one H100 at \$4.41 per hour. We scale measured H100 inference time
by the ratio of total input tokens to tokens processed on the H100,
accounting for requests completed on A40 GPUs. These estimates exclude
setup, model loading and idle time.

\end{samepage}

\begin{table}[t]
\centering\small
\setlength{\tabcolsep}{4pt}
\caption{API prices (USD per million tokens) and estimated total benchmark cost (USD). Cached denotes cache reads; cache writes cost 5 for Sol and 6.25 for Opus. A dash denotes no separate rate.}
\label{tab:api-costs}
\begin{tabular}{lrrrr}
\toprule
Model & Input & Cached & Output & Total \\
\midrule
Opus & 5 & 0.5 & 25 & 84.46 \\
Sol & 4 & 0.4 & 20 & 103.01 \\
Qwen & 2 & 0.2 & 6 & 13.30 \\
Kimi & 3 & 0.3 & 15 & 121.45 \\
DeepSeek & 1.74 & 0.348 & 3.48 & 9.32 \\
Gemma & 0.18 & 0.036 & 0.5 & 1.07 \\
Ministral & 0.2 & 0.04 & 0.2 & 1.70 \\
Nemotron & 0.2 & 0.04 & 0.6 & 3.81 \\
Jev & 0.042 & -- & 0 & 7.27 \\
\bottomrule
\end{tabular}
\end{table}

\begin{table}[t]
\centering\small
\setlength{\tabcolsep}{4pt}
\caption{GPU rental rates and estimated benchmark costs. Clef and Clef Flash use estimated H100 inference hours; other durations include setup and idle time. Thomson's cost is a lower bound.}
\label{tab:gpu-costs}
\begin{tabular}{lrrrr}
\toprule
Model & GPU & Hours & USD/h & Total (USD) \\
\midrule
Thomson & H100 & $\geq$ 6.186 & 4.41 & $\geq$ 27.28 \\
Snowdon & H100 & 3.598 & 4.41 & 15.87 \\
SaulLM & H200 & 2.390 & 4.47 & 10.68 \\
DISC-LawLLM & H100 & 1.653 & 4.41 & 7.29 \\
Kev & B300 & 1.665 & 8.00 & 13.32 \\
Clef & H100 & 5.732 & 4.41 & 25.28 \\
Clef Flash & H100 & 1.963 & 4.41 & 8.65 \\
\bottomrule
\end{tabular}
\end{table}

\section{Additional analyses}
\label{app:supporting}

\subsection{ALF selection volume}

Sol and Opus recover similar amounts of expert-marked
reasoning to DeepSeek, Snowdon and DISC-LawLLM while selecting
substantially fewer sentences (Table~\ref{tab:selection-volume}).
DeepSeek, for example, selects more
than twice the number of sentences for comparable recall. These
differences matter for assisted annotation: similar reasoning coverage
can leave an expert with considerably more highlighted text to check.

\begin{table*}[t]
\centering\small
\setlength{\tabcolsep}{4pt}
\caption{ALF precision and recall (\%), averaged over 471 opinions. Selections count scorable sentences across these opinions; the ratio divides this count by the 5,164 expert-selected sentences.}
\label{tab:selection-volume}
\label{tab:alf-metrics}
\begin{tabular}{lrrrr}
\toprule
Model & Precision & Recall & Selections & Ratio \\
\midrule
Opus & 47.1 & 75.5 & 9,150 & 1.77 \\
Sol & 48.3 & 75.8 & 9,295 & 1.80 \\
Qwen & 47.7 & 69.7 & 8,025 & 1.55 \\
Kimi & 31.4 & 87.8 & 18,155 & 3.52 \\
DeepSeek & 29.9 & 78.6 & 19,442 & 3.76 \\
Gemma & 57.4 & 53.7 & 4,966 & 0.96 \\
Thomson & 36.5 & 64.3 & 11,842 & 2.29 \\
Snowdon & 24.8 & 76.0 & 22,104 & 4.28 \\
SaulLM & 12.6 & 28.2 & 10,650 & 2.06 \\
Ministral & 17.1 & 84.2 & 35,082 & 6.79 \\
Nemotron & 16.2 & 83.6 & 35,010 & 6.78 \\
DISC-LawLLM & 14.1 & 76.7 & 33,250 & 6.44 \\
Kev & 47.5 & 65.5 & 7,860 & 1.52 \\
Jev & 27.9 & 79.3 & 18,560 & 3.59 \\
Clef & 41.5 & 67.4 & 9,129 & 1.77 \\
Clef Flash & 23.6 & 69.3 & 17,193 & 3.33 \\
\bottomrule
\end{tabular}
\end{table*}

\subsection{Decision models and joint LLM annotation}
\label{app:yes-no-comparison}

We compare Jev, Kev, Clef and Clef Flash with Gemma on the same 40
opinions to assess whether decision models offer advantages over
joint annotation by an LLM (Table~\ref{tab:gemma-control}). We also ask Gemma the same
yes/no questions to examine whether the comparison changes when it
uses the decision models' question format.

For the yes/no procedure, Gemma answers the Jev questions in separate
requests containing the full opinion and protocol. It uses the same
endpoint as joint selection, with temperature 0 and a 16-token output
cap, returning only \texttt{Yes} or \texttt{No}. All four decision
models use their saved outputs on these opinions at the fixed 0.5
cut-off.

None of the four decision models exceeds Gemma's joint selection in Macro-F1
on this sample. Kev nevertheless achieves higher F1 on ALF, Relevant
Precedents and Outcome, with lower scores on Procedural History and
Facts offsetting these gains. Jev and Clef are close to Gemma's joint
selection on Facts and Relevant Precedents, while Clef Flash's largest
gap is on ALF. The observed
advantages therefore depend on the category and decision model;
they do not amount to a general improvement over joint annotation.

Using separate yes/no questions changes the comparison: Jev, Kev and
Clef achieve higher Macro-F1 than Gemma, while Clef Flash remains
slightly below it. Gemma's yes/no procedure has the highest ALF F1.
Separate questions increase Gemma's recall and
reduce its precision across all four sentence categories, recovering
more reasoning but lowering F1 in every other category.
This comparison does not isolate architectural effects,
and Gemma's ALF F1 gain over joint selection remains uncertain, with
a paired 95\% bootstrap interval of [$-0.1$, 10.5] points, which
includes zero.

\begin{table*}[t]
\centering\small
\setlength{\tabcolsep}{4pt}
\caption{Exploratory comparison of joint selection and yes/no annotation on the same 40 opinions. Gemma (generative) uses both procedures; Jev, Kev, Clef and Clef Flash use yes/no questions with a 0.5 cut-off. F1 (\%) averages opinions with reference annotations: 40 for ALF and Facts, 36 for PH and Precedents, and 39 for Outcome.}
\label{tab:gemma-control}
\begin{tabular}{lrrrrrr}
\toprule
Model and procedure & ALF & PH & Precedents & Facts & Outcome & Macro-F1 \\
\midrule
Gemma (joint selection) & 47.5 & 53.0 & 60.5 & 77.0 & 79.6 & 63.5 \\
Gemma (yes/no) & 52.5 & 43.2 & 59.2 & 66.2 & 71.7 & 58.6 \\
Jev (yes/no) & 38.2 & 50.8 & 60.8 & 77.5 & 76.0 & 60.7 \\
Kev (yes/no) & 48.9 & 38.1 & 64.6 & 75.2 & 84.2 & 62.2 \\
Clef 27B (yes/no) & 45.4 & 42.0 & 62.1 & 77.6 & 73.7 & 60.2 \\
Clef Flash 9B (yes/no) & 32.1 & 45.9 & 58.3 & 77.0 & 77.5 & 58.2 \\
\bottomrule
\end{tabular}
\end{table*}

\begin{samepage}
\subsection{Thomson--Snowdon comparison}
\label{app:checkpoint-retention}

Thomson's improvement over Snowdon comes with more selective ALF
annotation. Under the same prompts and inference settings, Thomson
selects fewer sentences and achieves higher precision, but misses
more expert-marked reasoning. Its F1 improves by 7.3 points, with a
paired 95\% bootstrap interval of [5.6, 8.9]. An adapted model with
higher F1 can therefore still leave more reasoning for an annotator
to find.
Thomson's additional training includes non-legal material and further
post-training, so the improvement cannot be attributed to legal
adaptation alone.

\end{samepage}

\FloatBarrier
\section{Annotation protocol and model prompts}
\label{app:instructions}

\subsection{Expert annotation protocol}
\label{app:protocol}

Figure~\ref{fig:protocol-1} reproduces the original instructions used
by the human annotators, including iteration headings, amendments
and numbering. References to summaries, Restatement tags and
spreadsheet columns concern the original annotation project;
colours refer to its original annotation scheme.

\subsection{Model prompts}
\label{app:prompts}

Figures~\ref{fig:sentence-prompt-1} and~\ref{fig:precedent-prompt-1}
show the generative prompts for sentence selection and precedent
selection, with placeholders for the opinion text and candidate list.

\definecolor{instructionbackground}{gray}{0.96}
\newsavebox{\binaryquestionpanel}
\newenvironment{binaryquestion}{%
  \par\smallskip\begingroup
  \setlength{\fboxsep}{5pt}
  \begin{lrbox}{\binaryquestionpanel}
  \begin{minipage}{\dimexpr\columnwidth-2\fboxsep\relax}
  \normalfont\normalsize\raggedright
  \setlength{\parindent}{0pt}
}{%
  \end{minipage}
  \end{lrbox}
  \noindent\colorbox{instructionbackground}{\usebox{\binaryquestionpanel}}
  \par\endgroup\smallskip
}

\subsection{Binary questions}
\label{app:decision-questions}

All four decision models receive the full opinion and annotation instructions,
followed by a separate yes/no question for each sentence and category.
Each question asks whether a particular sentence belongs to one
annotation category. The example below, from \textit{Villanueva v.
Nowlin} (1966), asks whether sentence 59 belongs to Outcome.
The true and false criteria define the two possible answers.
\begin{binaryquestion}
\textbf{instructions:} Sentence [59] begins: ``The judgment is affirmed.''.
Following the instructions above, should sentence [59] be selected
as Outcome?
\par\smallskip
\textbf{true:} sentence [59] meets the Outcome definition above,
including its scope
\par\smallskip
\textbf{false:} sentence [59] does not meet the Outcome definition above
\end{binaryquestion}
For precedent selection, each question asks whether a case cited in
the opinion is relevant under the protocol. This example concerns
a candidate from the same opinion.
\begin{binaryquestion}
\textbf{instructions:} Following the instructions above, should
candidate [1] (Bradshaw v. Blystone Equipment Co.) be selected as a
relevant precedent?
\par\smallskip
\textbf{true:} candidate [1] meets the Relevant Precedents definition
above, including its scope
\par\smallskip
\textbf{false:} candidate [1] does not meet the Relevant Precedents
definition above
\end{binaryquestion}

Gemma receives the same questions and criteria, expressed as a request
to answer Yes or No. For the Outcome example above, the following text
appears after the full opinion and annotation instructions.
\begin{binaryquestion}
CURRENT DECISION\\
Sentence [59] begins: ``The judgment is affirmed.''.
Following the instructions above, should sentence [59] be selected
as Outcome?
\par\smallskip
\textbf{Yes:} sentence [59] meets the Outcome definition above,
including its scope\\
\textbf{No:} sentence [59] does not meet the Outcome definition above
\par\smallskip
Return only Yes or No.
\end{binaryquestion}
\newsavebox{\instructionpanel}

\begin{figure*}[tp]
\centering
\begingroup
\setlength{\fboxsep}{5pt}
\sbox{\instructionpanel}{\colorbox{instructionbackground}{%
\begin{minipage}{\dimexpr\textwidth-2\fboxsep\relax}
\normalsize\raggedright
\setlength{\parindent}{0pt}
\setlength{\parskip}{0pt}
\textbf{Protocols for Annotating Cases}
\par\vspace{1pt}
\textbf{Iteration \#5, 12 March 2021}
\par\vspace{1pt}
\textbf{Coding Protocols for Iteration \#4}
\par\vspace{1pt}
We will tag six types of text in the judgements, and add two elements.
\par\vspace{1pt}
\#1 Facts: these should be stated comprehensively early on in the judgement. Tag only text about facts that happened "in the world" before the litigation began (ie the facts alleged to have give rise to the legal consequences in the plaintiff's suit). Highlight these in yellow. Include any assumed/conceded "facts" as we are interested in all assumed facts on which the reasoning is based.
\par\vspace{1pt}
\#2 Procedural history: highlight in pink statements about the procedural status of the case ("this is an appeal ....") and ("plaintiff succeeded at first instance") and what standard of review is applied ("de novo" vs "review")  etc, and (where available) what the lower court's reasons for its decision were.
\par\vspace{1pt}
Turning to the actual decision in the case, we will focus on three specific aspects:
\par\vspace{1pt}
\#3 Application of law to facts: highlight in green the minimum necessary statements of the application of law to the facts to get the court to the decision. (Exclude discussion of points which are not necessary to the conclusion). Include contextual statements from the text that help to explain the application of the law to the facts. Include "judicial cues" for application of the law.
\par\vspace{1pt}
\#4 Relevant precedents -- highlight in blue the names of relevant prior cases (precedents) referred to which are salient for the application of law to facts in this case. The scope of this will be driven by the scope of what is tagged in green as "application of law to the facts". Include the citation information (court, year, law report).  This annotation can exclude cases cited by way of background, or in relation to issues which are not determinative. Please also include relevant pieces of legislation, including section numbers. (But no need to highlight statutory text)
\par\vspace{1pt}
\#5 Outcome: highlight in red at the end the court's statement of the outcome ("I find for the defendant..") Fine to highlight multiple times if it appears more than once.
\par\vspace{1pt}
In addition, we will also code a seventh factor, once we have completed the breakdown of rules of the restatement:
\par\vspace{1pt}
\#6 Plain English summary -- account in plain English of what happened in the case.
\par\vspace{1pt}
\#7 Restatement: indicate which rule(s) of the Restatement of Torts are applicable to the decision in the case. This will be based on a comparison of a table of Restatement provisions and our reading of the case.
\par\vspace{1pt}
\textbf{Additional Protocols Agreed During RA Meetings}
\par\vspace{1pt}
\begingroup\leftskip=1.5em\noindent\llap{1.\enspace}Our focus is on the following elements of tort law: existence/scope of duty, breach of duty (i.e., whether the defendant was negligent), and the existence of contributory negligence (i.e., whether the claimant was negligent) (the 'Relevant Issues').\par\endgroup
\par\vspace{1pt}
\begingroup\leftskip=1.5em\noindent\llap{2.\enspace}Pink and green highlights should only relate to the Relevant Issues. For example, if the appeal concerns (i) whether there was a duty; (ii) matters of causation; and (iii) damages, we should only highlight the procedural facts and any applications of the law to the case which relate to issue (i).\par\endgroup
\begingroup\leftskip=3.0em\noindent\llap{$\bullet$\enspace}Moreover, if the case primarily concerns issues (ii) and (iii), but issue (i) is mentioned very briefly (and what is mentioned would be useful to our model), annotate the case anyway, but mark it in the tagging sheet as a 'Marginal case' and add a comment to explain why you have marked it as such.\par\endgroup
\par
\end{minipage}}}
\typeout{PANEL:protocol:1:\the\dimexpr\ht\instructionpanel+\dp\instructionpanel\relax}
\usebox{\instructionpanel}
\endgroup
\caption{Expert annotation protocol.}
\label{fig:protocol-1}
\end{figure*}

\begin{figure*}[tp]
\ContinuedFloat
\centering
\begingroup
\setlength{\fboxsep}{5pt}
\sbox{\instructionpanel}{\colorbox{instructionbackground}{%
\begin{minipage}{\dimexpr\textwidth-2\fboxsep\relax}
\normalsize\raggedright
\setlength{\parindent}{0pt}
\setlength{\parskip}{0pt}
\begingroup\leftskip=1.5em\noindent\llap{3.\enspace}Contributory negligence:\par\endgroup
\begingroup\leftskip=3.0em\noindent\llap{$\bullet$\enspace}We are interested in whether the court finds the claimant to be guilty of contributory negligence. However, if the court goes on to discuss whether the claimant's contributory negligence caused his/her injury, then this does not need to be highlighted.\par\endgroup
\begingroup\leftskip=3.0em\noindent\llap{$\bullet$\enspace}Similarly, we can highlight by what percentage the court finds the claimant to have contributed to his/her injury, but we do not need to highlight any discussion of damage reduction on account of the claimant's negligence.\par\endgroup
\par\vspace{1pt}
\begingroup\leftskip=1.5em\noindent\llap{4.\enspace}Special liability rules:\par\endgroup
\begingroup\leftskip=3.0em\noindent\llap{$\bullet$\enspace}Where a case concerns occupiers' liability, employer's liability, strict liability, or other special rules governing liability, the discussions relating to these issues need only be highlighted if they are generalizable to common law negligence. For example:\par\endgroup
\begingroup\leftskip=4.5em\noindent\llap{(a)\enspace}If the case exclusively concerns an employee's attempt to claim under the New Mexican Workers' Compensation Act, then this should not be highlighted.\par\endgroup
\begingroup\leftskip=4.5em\noindent\llap{(b)\enspace}If, however, the case discusses an occupier's duty of care to visitors and does so in general tort law of negligence terms, this should be highlighted (even if the case refers to some specific rules under the Restatement).\par\endgroup
\begingroup\leftskip=3.0em\noindent\llap{$\bullet$\enspace}In the case (b), we should mark the column 'Potentially other sections of the Restatement relevant' in the tagging sheet.\par\endgroup
\par\vspace{1pt}
\begingroup\leftskip=1.5em\noindent\llap{5.\enspace}If the claim is against a state entity, and the discussion of the court merely concerns whether that state entity is immune from liability under the Tort Claims Act, this should not be highlighted.\par\endgroup
\par\vspace{1pt}
\begingroup\leftskip=1.5em\noindent\llap{6.\enspace}We do not highlight the cases related to the standard of review.\par\endgroup
\par\vspace{1pt}
\begingroup\leftskip=1.5em\noindent\llap{7.\enspace}If the court makes it clear that through its judgment it is making 'new law', mark the 'Makes new law' column in the tagging sheet.\par\endgroup
\par\vspace{1pt}
\begingroup\leftskip=1.5em\noindent\llap{8.\enspace}Mark the column 'special concurrence' in the tagging sheet if the case includes a special concurrence after the judgment. The contents of the special concurrence should not be annotated.\par\endgroup
\par\vspace{1pt}
\begingroup\leftskip=1.5em\noindent\llap{9.\enspace}Do not annotate dissenting opinions.\par\endgroup
\par\vspace{1pt}
\begingroup\leftskip=1.5em\noindent\llap{10.\enspace}Treat witness testimony which is relied on by the court for its reasoning as fact.\par\endgroup
\begingroup\leftskip=3.0em\noindent\llap{$\bullet$\enspace}Where some witnesses disagree, but the court makes it clear that a majority of witnesses agree on particular facts, do not treat 'minority' witness testimony as fact.\par\endgroup
\par\vspace{1pt}
\begingroup\leftskip=1.5em\noindent\llap{11.\enspace}Highlight the whole case name, including all the citation references after it (so the blue highlights will sometimes be very long).\par\endgroup
\par\vspace{1pt}
\begingroup\leftskip=1.5em\noindent\llap{12.\enspace}The various tort law Restatements should be highlighted blue even though they are not technically statutory.\par\endgroup
\par\vspace{1pt}
\begingroup\leftskip=1.5em\noindent\llap{13.\enspace}If the court mentions claims made by the claimant against parties that are not party to the case at hand, this information does not form part of the procedural facts and therefore should not be highlighted pink.\par\endgroup
\par\vspace{1pt}
\begingroup\leftskip=1.5em\noindent\llap{14.\enspace}Square brackets and other characters/symbols can be deleted or ignored (the model will delete them anyway).\par\endgroup
\par\vspace{1pt}
\begingroup\leftskip=1.5em\noindent\llap{15.\enspace}Add any other comments in the 'Comments' section of the tagging sheet that you think are necessary.\par\endgroup
\par\vspace{1pt}
\begingroup\leftskip=1.5em\noindent\llap{16.\enspace}Where new factual information is recounted by the court in the course of its application of the law to the facts, tag that information as (yellow) fact notwithstanding that it is part of the court's reasoning, to ensure that all facts are captured.\par\endgroup
\par\vspace{1pt}
\begingroup\leftskip=1.5em\noindent\llap{17.\enspace}Where the judgement includes a detailed factual description of the life, behaviour or practices of one of the parties (such as psychiatric or employment history) that occurs prior to the factual situation in which the tortious acts are alleged to have occurred, this should be annotated as (yellow) fact.\par\endgroup
\par
\end{minipage}}}
\typeout{PANEL:protocol:2:\the\dimexpr\ht\instructionpanel+\dp\instructionpanel\relax}
\usebox{\instructionpanel}
\endgroup
\caption{Expert annotation protocol (continued).}
\label{fig:protocol-2}
\end{figure*}

\begin{figure*}[tp]
\ContinuedFloat
\centering
\begingroup
\setlength{\fboxsep}{5pt}
\sbox{\instructionpanel}{\colorbox{instructionbackground}{%
\begin{minipage}{\dimexpr\textwidth-2\fboxsep\relax}
\normalsize\raggedright
\setlength{\parindent}{0pt}
\setlength{\parskip}{0pt}
\begingroup\leftskip=1.5em\noindent\llap{18.\enspace}Where facts are recounted by the court that pertain exclusively to issues that are out of scope, they should nonetheless be annotated.\par\endgroup
\begingroup\leftskip=3.0em\noindent\llap{$\bullet$\enspace}For example, see the case of Trujillo v Chavez. In this case, the court recounts facts that relate exclusively to the question of whether or not the plaintiff was a fare-paying passenger (in which case the plaintiff would not be statutorily barred from seeking compensation). Notwithstanding that those facts do not go to the negligence of the defendant, they should be annotated.\par\endgroup
\par\vspace{1pt}
\begingroup\leftskip=1.5em\noindent\llap{19.\enspace}Where a case is heavily based on State-specific law (for example Delgado Claims in New Mexico), it should still be tagged if it is a duty and breach case.\par\endgroup
\par\vspace{1pt}
\begingroup\leftskip=1.5em\noindent\llap{20.\enspace}Where the court's statement of facts takes the form of (a) restatement of witness testimony (apparently uncontradicted, and ostensibly in support of conclusions that such testimony establishes the facts); or (b) a claim that 'the evidence establishes' the facts recounted, exclude prefatory remarks that refer to the basis for the court's conclusion such as, 'The evidence establishes that ...', and 'The plaintiff testified that ...'.\par\endgroup
\begingroup\leftskip=3.0em\noindent\llap{$\bullet$\enspace}Where the court's statement of facts includes an excerpt from examination-in-chief or cross-examination of a witness, the useful, unambiguous portions of those excerpts (that is, excluding exclamations such as, 'oh' and 'um' which accompany verbatim records of witness examination) should generally be annotated as (yellow) fact, EXCEPT where necessary to give proper meaning to the court's application of law to the facts on the point in support of which the excerpt is included, in which case such testimony should be highlighted as (green) application of the law to the facts.\par\endgroup
\par\vspace{1pt}
\begingroup\leftskip=1.5em\noindent\llap{21.\enspace}Where the court's application of law to the facts refers to a relevant case name or statute, highlight that case name or statute in blue (keeping the rest of the statement in green).\par\endgroup
\begingroup\leftskip=3.0em\noindent\llap{$\bullet$\enspace}If, however, the procedural facts refer to a case name or statute, do not highlight this blue: the procedural facts should be highlighted pink only.\par\endgroup
\par\vspace{1pt}
\begingroup\leftskip=1.5em\noindent\llap{22.\enspace}Where the court states that a piece of evidence included in an affidavit and excerpted in the judgement is not admissible, and therefore that it does not form a basis for the court's decision, that evidence should not be highlighted as fact.\par\endgroup
\begingroup\leftskip=3.0em\noindent\llap{$\bullet$\enspace}EXAMPLE: Consider the case of Cowan v. Ellison Enterprises, Inc., in which the court states the following in relation to part of the plaintiff's affidavit:\par\endgroup
\par\vspace{1pt}
In reviewing the motion for summary judgment, we do not consider appellant's statement, "friends and relatives . . . have seen loose grapes on the floor in the produce section." That statement was not based on personal knowledge, and it is nothing more than inadmissible hearsay. As such, it should not be accepted as the basis for finding a genuine issue of material fact to deny entry of summary judgment.
\par\vspace{1pt}
\begingroup\leftskip=1.5em\noindent\llap{23.\enspace}Where the court discusses whether a particular piece of evidence has probative value as regards the tort elements of duty or breach, and the discussion provides a signal as to the court's reasoning on those elements, this discussion should be annotated.\par\endgroup
\par\vspace{1pt}
\begingroup\leftskip=1.5em\noindent\llap{24.\enspace}Where facts appear in a dissenting opinion, those facts should ordinarily not be annotated unless there is no effective statement of the facts in the majority opinion.\par\endgroup
\par\vspace{1pt}
\begingroup\leftskip=1.5em\noindent\llap{25.\enspace}Where facts appear in a specially concurring separate opinion that do not appear in the majority opinion, they should be annotated.\par\endgroup
\par\vspace{1pt}
\begingroup\leftskip=1.5em\noindent\llap{26.\enspace}Cases dealing with damage to property should be annotated where they turn on the same elements of negligence considered for the purposes of cases involving physical injury and they do not concern pure economic loss.\par\endgroup
\par\vspace{1pt}
\begingroup\leftskip=1.5em\noindent\llap{27.\enspace}Where possible, do not annotate references to the party type (such as 'defendant', 'plaintiff', or 'appellant') in the court's restatement of the facts.\par\endgroup
\begingroup\leftskip=3.0em\noindent\llap{$\bullet$\enspace}For example: if the court refers to 'defendant John' in its statement of the facts, highlight only 'John' in yellow.\par\endgroup
\par
\end{minipage}}}
\typeout{PANEL:protocol:3:\the\dimexpr\ht\instructionpanel+\dp\instructionpanel\relax}
\usebox{\instructionpanel}
\endgroup
\caption{Expert annotation protocol (continued).}
\label{fig:protocol-3}
\end{figure*}
\begin{figure*}[tp]
\centering
\begingroup
\setlength{\fboxsep}{5pt}
\sbox{\instructionpanel}{\colorbox{instructionbackground}{%
\begin{minipage}{\dimexpr\textwidth-2\fboxsep\relax}
\normalsize\raggedright
\setlength{\parindent}{0pt}
\setlength{\parskip}{0pt}
You are annotating a judicial opinion in a civil negligence case.\par
\par\vspace{2pt}
Read the entire supplied opinion. Each sentence is prefixed with its integer ID in square brackets. Select the sentences belonging to these four categories:\par
\begingroup\leftskip=1.3em\noindent\llap{-\enspace}Facts\par\endgroup
\begingroup\leftskip=1.3em\noindent\llap{-\enspace}Procedural History\par\endgroup
\begingroup\leftskip=1.3em\noindent\llap{-\enspace}Application of Law to Facts\par\endgroup
\begingroup\leftskip=1.3em\noindent\llap{-\enspace}Outcome\par\endgroup
\par\vspace{2pt}
Select every category that applies to a sentence. Different portions of one sentence may qualify for different categories. Leave sentences that do not qualify for any category unlabeled. Base your annotations only on the supplied opinion. Do not use external sources or infer unstated facts.\par
\par\vspace{2pt}
\textbf{SCOPE}
\par\vspace{2pt}
The Relevant Issues are:\par
\begingroup\leftskip=1.3em\noindent\llap{1.\enspace}The existence or scope of a duty of care.\par\endgroup
\begingroup\leftskip=1.3em\noindent\llap{2.\enspace}Breach of that duty: whether the defendant was negligent.\par\endgroup
\begingroup\leftskip=1.3em\noindent\llap{3.\enspace}Contributory negligence: whether the claimant was negligent.\par\endgroup
\par\vspace{2pt}
Procedural History and Application of Law to Facts are restricted to these Relevant Issues. These restrictions do not exclude Facts about other issues or the court's Outcome.\par
\par\vspace{2pt}
Discussion solely about causation or damages is outside that scope. Discussion that merely concerns whether a state entity is immune under a Tort Claims Act is also outside that scope.\par
\par\vspace{2pt}
Special liability regimes, including occupiers' liability, employers' liability, strict liability, and workers' compensation, are within scope only where the court reasons in terms generalizable to common-law negligence. State-specific law does not by itself exclude a passage: include it where it concerns duty or breach.\par
\par\vspace{2pt}
Property-damage cases are within scope where they turn on the same negligence elements as physical-injury cases and do not concern pure economic loss.\par
\par\vspace{2pt}
A passage addressing a Relevant Issue together with an excluded issue is not excluded merely because it mentions the excluded issue.\par
\par\vspace{2pt}
Annotate only the supplied majority opinion. Do not annotate a dissent or separate concurrence.\par
\par\vspace{2pt}
\textbf{FACTS}
\par\vspace{2pt}
Events and circumstances that happened before the litigation began, including:\par
\begingroup\leftskip=1.3em\noindent\llap{-\enspace}The events alleged to give rise to the claim.\par\endgroup
\begingroup\leftskip=1.3em\noindent\llap{-\enspace}Facts accepted or assumed by the court for its reasoning, including conceded facts; they need not have been conclusively established.\par\endgroup
\begingroup\leftskip=1.3em\noindent\llap{-\enspace}Background about a party's history, behavior, or practices.\par\endgroup
\begingroup\leftskip=1.3em\noindent\llap{-\enspace}Facts concerning issues outside the Relevant Issues.\par\endgroup
\begingroup\leftskip=1.3em\noindent\llap{-\enspace}New factual information recounted within the court's reasoning.\par\endgroup
\par\vspace{2pt}
Annotate Facts comprehensively, wherever they appear.\par
\par\vspace{2pt}
Witness testimony relied on by the court is treated as Facts. Where the court recounts testimony or states that "the evidence establishes" something, the recounted content is Facts; the introductory phrase is not separately relevant.\par
\par\vspace{2pt}
Where witnesses disagree and the court clearly relies on the account supported by the majority of witnesses, do not annotate the conflicting minority account as Facts.\par
\par\vspace{2pt}
Do not annotate evidence that the court rules inadmissible and does not rely on as Facts. The testimony exception in the Application of Law to Facts definition below takes precedence where it applies.\par
\par
\end{minipage}}}
\typeout{PANEL:sentence-annotation:1:\the\dimexpr\ht\instructionpanel+\dp\instructionpanel\relax}
\usebox{\instructionpanel}
\endgroup
\caption{Model prompt for sentence annotation.}
\label{fig:sentence-prompt-1}
\end{figure*}

\begin{figure*}[tp]
\ContinuedFloat
\centering
\begingroup
\setlength{\fboxsep}{5pt}
\sbox{\instructionpanel}{\colorbox{instructionbackground}{%
\begin{minipage}{\dimexpr\textwidth-2\fboxsep\relax}
\normalsize\raggedright
\setlength{\parindent}{0pt}
\setlength{\parskip}{0pt}
\textbf{PROCEDURAL HISTORY}
\par\vspace{2pt}
Statements about:\par
\begingroup\leftskip=1.3em\noindent\llap{-\enspace}The procedural status of the case.\par\endgroup
\begingroup\leftskip=1.3em\noindent\llap{-\enspace}What happened in the proceedings below.\par\endgroup
\begingroup\leftskip=1.3em\noindent\llap{-\enspace}The lower court's reasons for its decision, where given.\par\endgroup
\begingroup\leftskip=1.3em\noindent\llap{-\enspace}The standard of review applied by the present court.\par\endgroup
\par\vspace{2pt}
Select these only where they relate to a Relevant Issue.\par
\par\vspace{2pt}
Descriptions of claims against persons who are not parties to the present case are not Procedural History for this task.\par
\par\vspace{2pt}
\textbf{APPLICATION OF LAW TO FACTS}
\par\vspace{2pt}
Application of Law to Facts consists of the minimum necessary statements of the court's application of law to facts to reach its decision on a Relevant Issue. It excludes discussion of points not necessary to the conclusion.\par
\par\vspace{2pt}
It includes contextual statements that help explain the application of law to facts, and judicial cues indicating that the court is applying the law.\par
\par\vspace{2pt}
It includes discussion of whether evidence has probative value on duty or breach when that discussion signals the court's reasoning on those issues.\par
\par\vspace{2pt}
Where an excerpt of witness testimony is necessary to give proper meaning to the court's application of law to facts on a point, that excerpt belongs to Application of Law to Facts rather than Facts.\par
\par\vspace{2pt}
For contributory negligence, it includes the court's determination of whether the claimant was negligent and any percentage of contribution. It excludes discussion solely about whether the claimant's negligence caused the injury or how damages should be reduced.\par
\par\vspace{2pt}
Merely reporting a party's argument does not by itself constitute Application of Law to Facts. Such material qualifies only where it forms part of the court's own analysis or provides context necessary to explain that analysis.\par
\par\vspace{2pt}
\textbf{OUTCOME}
\par\vspace{2pt}
The present court's explicit statement of its outcome, such as finding for a party, affirming, reversing, remanding, dismissing, modifying, or otherwise disposing of the case or a claim.\par
\par\vspace{2pt}
Select the court's statement of the outcome. You may select multiple statements where the outcome appears more than once.\par
\par\vspace{2pt}
A description of the lower court's disposition belongs to Procedural History when that definition applies.\par
\par\vspace{2pt}
\textbf{OUTPUT}
\par\vspace{2pt}
Return only a valid JSON object with exactly these four keys:\par
\begin{tabular}{@{}l@{}}
\texttt{\{}\\
\texttt{\quad "Facts": [],}\\
\texttt{\quad "Procedural History": [],}\\
\texttt{\quad "Application of Law to Facts": [],}\\
\texttt{\quad "Outcome": []}\\
\texttt{\}}
\end{tabular}
\par\vspace{2pt}
Each array must contain the integer IDs of the selected sentences in ascending order. Use only IDs present in the supplied opinion. Do not repeat an ID within an array. An ID may appear in multiple category arrays when the sentence qualifies for each category. Use an empty array when no sentence qualifies for a category. Do not include explanations, quotations, or Markdown formatting.\par
\par\vspace{2pt}
\textbf{OPINION}
\par\vspace{2pt}
\texttt{\{\{NUMBERED\_SENTENCES\}\}}
\par
\end{minipage}}}
\typeout{PANEL:sentence-annotation:2:\the\dimexpr\ht\instructionpanel+\dp\instructionpanel\relax}
\usebox{\instructionpanel}
\endgroup
\caption{Model prompt for sentence annotation (continued).}
\label{fig:sentence-prompt-2}
\end{figure*}

\begin{figure*}[tp]
\centering
\begingroup
\setlength{\fboxsep}{5pt}
\sbox{\instructionpanel}{\colorbox{instructionbackground}{%
\begin{minipage}{\dimexpr\textwidth-2\fboxsep\relax}
\small\raggedright
\setlength{\parindent}{0pt}
\setlength{\parskip}{0pt}
You are identifying relevant precedents in a judicial opinion in a civil negligence case.\par
\par\vspace{2pt}
Read the entire supplied opinion and select the relevant prior cases from the numbered candidate list.\par
\par\vspace{2pt}
Base your selection only on how the supplied opinion uses each case. Do not use external sources or select a case merely because you recognize its name.\par
\par\vspace{2pt}
\textbf{SCOPE}
\par\vspace{2pt}
The Relevant Issues are:\par
\begingroup\leftskip=1.3em\noindent\llap{1.\enspace}The existence or scope of a duty of care.\par\endgroup
\begingroup\leftskip=1.3em\noindent\llap{2.\enspace}Breach of that duty: whether the defendant was negligent.\par\endgroup
\begingroup\leftskip=1.3em\noindent\llap{3.\enspace}Contributory negligence: whether the claimant was negligent.\par\endgroup
\par\vspace{2pt}
Procedural History, Application of Law to Facts, and Relevant Precedents are restricted to these Relevant Issues. These restrictions do not exclude Facts about other issues or the court's Outcome.\par
\par\vspace{2pt}
Discussion solely about causation or damages is outside that scope. Discussion that merely concerns whether a state entity is immune under a Tort Claims Act is also outside that scope.\par
\par\vspace{2pt}
Special liability regimes, including occupiers' liability, employers' liability, strict liability, and workers' compensation, are within scope only where the court reasons in terms generalizable to common-law negligence. State-specific law does not by itself exclude a passage: include it where it concerns duty or breach.\par
\par\vspace{2pt}
Property-damage cases are within scope where they turn on the same negligence elements as physical-injury cases and do not concern pure economic loss.\par
\par\vspace{2pt}
A passage addressing a Relevant Issue together with an excluded issue is not excluded merely because it mentions the excluded issue.\par
\par\vspace{2pt}
Annotate only the supplied majority opinion. Do not annotate a dissent or separate concurrence.\par
\par\vspace{2pt}
\textbf{APPLICATION OF LAW TO FACTS}
\par\vspace{2pt}
Application of Law to Facts consists of the minimum necessary statements of the court's application of law to facts to reach its decision on a Relevant Issue. It excludes discussion of points not necessary to the conclusion.\par
\par\vspace{2pt}
It includes contextual statements that help explain the application of law to facts, and judicial cues indicating that the court is applying the law.\par
\par\vspace{2pt}
It includes discussion of whether evidence has probative value on duty or breach when that discussion signals the court's reasoning on those issues.\par
\par\vspace{2pt}
Where an excerpt of witness testimony is necessary to give proper meaning to the court's application of law to facts on a point, that excerpt belongs to Application of Law to Facts rather than Facts.\par
\par\vspace{2pt}
For contributory negligence, it includes the court's determination of whether the claimant was negligent and any percentage of contribution. It excludes discussion solely about whether the claimant's negligence caused the injury or how damages should be reduced.\par
\par\vspace{2pt}
Merely reporting a party's argument does not by itself constitute Application of Law to Facts. Such material qualifies only where it forms part of the court's own analysis or provides context necessary to explain that analysis.\par
\par\vspace{2pt}
\textbf{RELEVANT PRECEDENTS}
\par\vspace{2pt}
The relevant precedents are the prior cases referred to by the court that are salient to its Application of Law to Facts in this opinion, as defined above.\par
\par\vspace{2pt}
The scope of Relevant Precedents follows the scope of Application of Law to Facts.\par
\par\vspace{2pt}
Exclude cases cited only:\par
\begingroup\leftskip=1.3em\noindent\llap{-\enspace}By way of background.\par\endgroup
\begingroup\leftskip=1.3em\noindent\llap{-\enspace}For the standard of review.\par\endgroup
\begingroup\leftskip=1.3em\noindent\llap{-\enspace}In relation to issues that are not determinative.\par\endgroup
\begingroup\leftskip=1.3em\noindent\llap{-\enspace}In the procedural history: a case named in the procedural recital is not thereby a relevant precedent.\par\endgroup
\par\vspace{2pt}
Consider all uses of a candidate throughout the opinion. A case cited for an excluded purpose in one passage may still qualify if it is salient to the court's Application of Law to Facts elsewhere.\par
\par\vspace{2pt}
Each candidate represents one distinct prior case. Full citations, short-form references, and attributable "id." or "supra" references to that case belong to the same candidate.\par
\par\vspace{2pt}
This task covers prior cases only. Do not select or add statutes, regulations, Restatement sections, or treatises.\par
\par\vspace{2pt}
\textbf{OUTPUT}
\par\vspace{2pt}
Return only a valid JSON object with exactly this key:\par
\begin{tabular}{@{}l@{}}
\texttt{\{}\\
\texttt{\quad "relevant": []}\\
\texttt{\}}
\end{tabular}
\par\vspace{2pt}
The array must contain the integer IDs of the selected candidates in ascending order. Use only IDs present in the supplied candidate list. Do not repeat IDs. Use an empty array when no candidate qualifies. Do not include sentence annotations, relevance scores, explanations, quotations, or Markdown formatting.\par
\par\vspace{2pt}
\textbf{OPINION}
\par\vspace{2pt}
\texttt{\{\{CLEANED\_OPINION\}\}}
\par\vspace{2pt}
\textbf{CANDIDATE CASES}
\par\vspace{2pt}
\texttt{\{\{NUMBERED\_CANDIDATES\}\}}
\par
\end{minipage}}}
\typeout{PANEL:precedent-selection:1:\the\dimexpr\ht\instructionpanel+\dp\instructionpanel\relax}
\usebox{\instructionpanel}
\endgroup
\caption{Model prompt for precedent selection.}
\label{fig:precedent-prompt-1}
\end{figure*}